\documentclass[11pt]{article}
\usepackage{coling2020}
\usepackage{times}
\usepackage{url}
\usepackage{latexsym}
\usepackage{graphicx}
\usepackage{subcaption}
\usepackage{tabularx}
\usepackage[export]{adjustbox}
\usepackage{soul}
\usepackage{arabtex}
\usepackage{multirow}
\usepackage{makecell}
\usepackage{utf8}
\usepackage{multirow}

\usepackage{enumitem,kantlipsum}

\usepackage{ulem}

\usepackage{wrapfig,lipsum,booktabs}

\usepackage{epstopdf}
\usepackage[latin1]{inputenc}

\usepackage{hyperref}
\usepackage{xstring}
\usepackage{soul}
\usepackage{color}
\setcode{utf8}

\newcommand{\QADI}{QADI }
\newcommand{\MADAR}{MADAR }

\colingfinalcopy 

\title{Arabic Dialect Identification in the Wild}

\author{Ahmed Abdelali, Hamdy Mubarak, Younes Samih, Sabit Hassan and Kareem Darwish \\
  Qatar Computing Research Institute \\
  Hamad Bin Khalifa University (HBKU) \\
 Doha, Qatar \\
  {\tt \{aabdelali,hmubarak,ysamih,sahassan2,kdarwish\}@hbku.edu.qa} }

\date{}

\begin{document}
\maketitle
\begin{abstract}
We present QADI, an automatically collected dataset of tweets belonging to a wide range of country-level Arabic dialects \textemdash covering 18 different countries in the Middle East and North Africa region. 
Our method for building this dataset relies on applying multiple filters to identify users who belong to different countries based on their account descriptions and to eliminate tweets that are either written in Modern Standard Arabic or contain inappropriate language.  The resultant dataset contains 540k tweets from 2,525 users who are evenly 
distributed across 18 Arab countries. 
Using intrinsic evaluation, we show that the labels of a set of randomly selected tweets are 91.5\% accurate.  For extrinsic evaluation, we are able to build effective country-level dialect identification on tweets with a macro-averaged F1-score of 60.6\% across 18 classes.
\end{list} 
\end{abstract}

\section{Introduction}
\begin{figure}[!h]
\begin{center}
\includegraphics[scale=0.8]{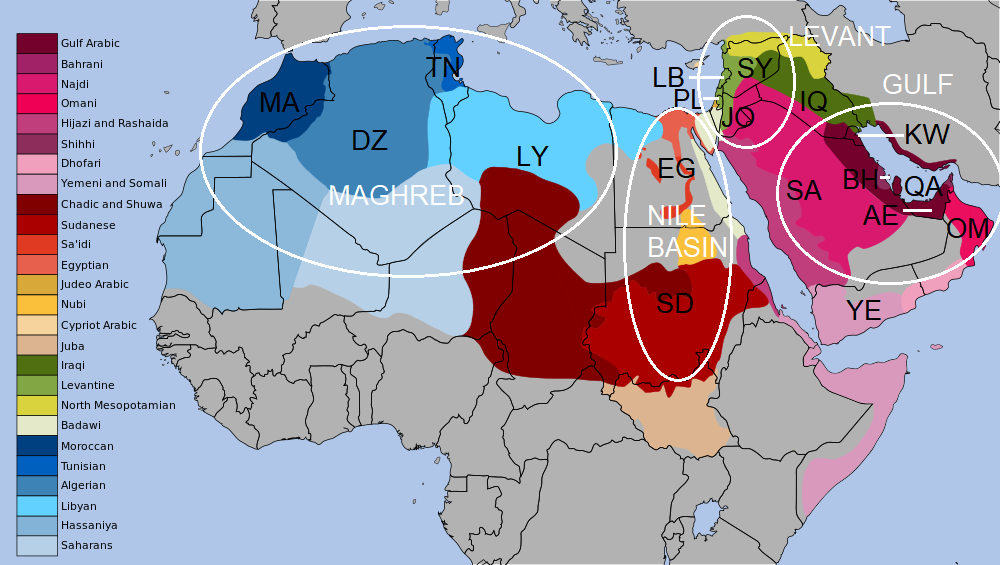} 
\caption{Geographical distribution of Arabic dialects. (Source: \url{https://en.wikipedia.org/wiki/Varieties_of_Arabic}, country codes and regions are added).}
\label{fig:arabic-dialects-map}
\end{center}
\end{figure}

Twitter is one of the most popular social media platforms in the Middle East and North Africa (MENA) region with almost two thirds (63\%) of Arab youth indicating that they look first to Facebook and Twitter for news \cite{radcliffe2019state}.  The popularity of Twitter in MENA is reflected by approximately 164 million active monthly users, who produce a massive volume of Arabic tweets, much of which is in Dialectal Arabic (DA).  
Hence, many researchers have been using Twitter as a major data source that is representative of current language and linguistic phenomena \cite{mubarak2014using,samih-etal-2017-learning,ZaghouaniCharfi2018}. Though Arabic is the lingua franca of most of the MENA region, different dialects of Arabic are used in different countries. While some dialects may differ significantly from each other (e.g. Egyptian dialect (EG) and Moroccan Maghrebi dialect (MA)\footnote{We use ISO 3166-1 alpha-2 for country codes:  \url{https://en.wikipedia.org/wiki/List\_of\_ISO\_3166\_country\_codes}}), others, particularly those in close in geographic proximity, may be more difficult to tweak apart (e.g. variants of the Levantine dialect such as Syrian (SY) and Lebanese (LB)). Figure~\ref{fig:arabic-dialects-map} highlights the dialectal variations across the Arabic world. The figure shows that dialects are a continuum that often transcends geographical regions and boarders.  Automatically distinguishing between the different dialectal variations is valuable for many downstream applications such as machine translations \cite{Diab:2014:tharwa}, POS tagging \cite{darwish2020effective}, geo-locating users, and author profiling \cite{Sadat:2014:automatic}.

Though there has been prior work on performing Arabic Dialect Identification (ADI), much of the work was conducted on datasets with significant limitations in terms of genre \cite{Bouamor:2018:madar,Zaidan:2011:arabic}, even coverage of different dialects, or focus~\cite{bouamor-etal-2019-madar}, where the focus is on geo-locating users as opposed to identifying dialects. In this work, we 
expand beyond these efforts by utilizing tweets from across the MENA region to build a large non-genre specific fine-grained and balanced country-level dialectal Arabic dataset that we use to build effective Arabic Dialect Identification. 

We rely on two main features to build the dataset.  The first feature is the Twitter user profile description, where we identify users who self-declare themselves as belonging to a specific country in different forms such as showing signs of loyalty and pride (e.g. ``proud Egyptian''). In the second, we use a classifier that utilizes distant supervision to accurately discriminates between MSA and dialects.  
In doing so, we can identify users who self-declare their identity, mostly tweet in dialectal Arabic,  
and only retain dialectal user tweets. 
Further, we use our newly constructed dataset to build models that can effectively distinguish between 18 country-level Arabic dialects. We didn't consider four Arab countries (namely Mauritania, Somalia, Djibouti and Comoros), because we were not able to find a sufficient number of Twitter users.   
This could be due to the limited use of Twitter in these countries, or that users may tweet primarily in other languages. 
For automated dialect identification, our models use a variety of features, such as character- and word-level n-gram, static word embeddings, 
and contextual embeddings (e.g. BERT$_{base-multilingual}$ and AraBERT), and two classification techniques, namely Support Vector Machines (SVM) classification and fine-tuned Transformer models. 

The contributions of this work are:
\begin{itemize}
    \item We introduce a new method for constructing a highly accurate Arabic dialectal dataset from Twitter. This method can be completely automated such that it can be used in the future to collect fresh dialectal tweets.
    \item We build the QCRI 
    Arabic Dialects Identification (QADI)\footnote{QADI Dataset, which make freely available for
research purposes from
\url{http://alt.qcri.org/resources/qadi/}}. It is the largest balanced non-genre specific country-level Arabic dialectal tweet dataset.  The dataset contains more than 540k tweets covering 18 dialects with an associated test set containing 182 tweets per country on average that was manually labeled by native speakers from 18 Arab countries.  
    \item We provide a list of Twitter accounts from 18 Arab countries  (a total of 2,525 accounts with an average of 140 accounts per country) that can be used in author profiling tasks.
    \item We use the new dataset to build state-of-the-art tweet-level Arabic dialect identification models using a variety of features and classifiers.  
\end{itemize}

\section{Related Work}

Most efforts in building resources for Arabic dialect identification are limited either in terms of the genre, granularity, or the size of the data. Zaidan and Callison-Burch \shortcite{Zaidan:2011:arabic} curated the Arabic Online Commentary Dataset, a resource of more than 52M-words. They annotated over 108K sentences (41\%) of the dataset with one of 5 possible dialects, namely: Maghrebi, Egyptian, Levantine, Gulf, and Iraqi. Similarly Alshutayri and Atwell~\shortcite{Alshutayrietal2017} and El-Haj et al.~\shortcite{el-haj-etal-2018-arabic} annotated collections of texts using the five regions/dialects. Elfardy and Diab  \shortcite{Elfardy:2013:sentence} and Darwish et al. \shortcite{Darwish:2014:verifiably} identified whether a sentence is Modern Standard Arabic (MSA) or Egyptian.

In recent years, there have been more efforts to cover more countries with finer granularity. Abdul-Mageed et al. \shortcite{Abdul-Mageed:2018:you} constructed a dataset that covers 10 countries. Zaghouani and Charfi~\shortcite{ZaghouaniCharfi2018} built the ``Arap-Tweet'' dataset, which includes tweets from 16 countries.  The MADAR (Multi-Arabic Dialect Applications and Resources) project ~\cite{Bouamor:2018:madar,bouamor-etal-2019-madar} produced several resources among which two corpora were used for the shared task of fine-grained dialect identification~\cite{bouamor-etal-2019-madar}. The resources includes a lexicon and a 1,000 parallel sentences from the travel domain that were translated into local dialects of 26 Arab cities. Additionally, the project released another set of tweets by searching twitter using a set of 25 seed hashtags corresponding to the 22 states of the Arab League (e.g., \#Algeria, \#Egypt, \#Kuwait, etc.) and relevant hashtags such as: ``\#ArabWorld'', ``\#ArabLeague'', and ``\#Arab''. The approach resulted in collecting over 2,980 profiles.  When inspecting the profiles, The majority of the obtained users were from Saudi Arabia, representing 36\% of the total profiles. 
This was another motivation to curate a more balanced and representative dataset to use for dialect identification. Further, as we show later, this dataset is 
sub-optimal for  
tweet-level dialect identification.

Multiple approaches have been used for dialect ID that exploit a variety of features, such as character or word n-grams~\cite{Darwish:2014:verifiably,zaidan:2014:arabic,malmasi-etal-2016-discriminating,Sadat:2014:automatic}, and techniques such as multiple kernel learning ~\cite{ionescu-popescu-2016-unibuckernel} and  distributed representation of dialects~ \cite{Abdul-Mageed:2018:you} to name a few. Zhang and Abdul-Mageed \shortcite{no-army} used semi-supervised learning using BERT$_{base-multilingual}$ model for user-level dialect identification on the MADAR Shared Task. 


\section{Data Collection}
It is common for users on social networks to disclose 
social and linguistic information about themselves in their profiles. In Twitter, the user profile provides a header and a short biography. Both fields allow users to freely  describe themselves. Surveying Arabic speaking profiles, it is customary to see many users declaring their patriotism and national belonging by using their county's flag or explicitly naming the city or country that they are from (e.g. `Kuwait is my home country'', ``Libyan citizen'').  Table  \ref{tab:ExamplesOfSelfDeclaration} shows some examples of such self-disclosures.  
To build our dataset, we obtained a collection of Arabic tweets that was crawled using the Twitter streaming API, where we set the language filter to Arabic ("lang:ar"), during the entirety of March and April, 2018. In all, the collection contains 25M tweets from which we extracted the profile information of all the users who authored these tweets. We applied three filters on user profiles and tweets as we describe in the next subsections.  

\begin{table}[]
    \centering
    \begin{tabular}{r|l|l}
        Arabic & Transliteration & Translation \\ \hline
        \<سعودي وأفتخر>
         & sEwdy w$>$ftKr & Saudi and proud\\
        \<بنت دمشق>
         & bnt dm\$q & the daughter of Damascus \\
        \<أنا من تونس>
         & $>$nA mn twns & I am from Tunisia \\ 
    \end{tabular}
    \caption{Examples of user self-disclosure in Twitter profile descriptions.}
    \label{tab:ExamplesOfSelfDeclaration}
\end{table}

\subsection{Country Identification}
For the first stage, to identify the user's country, we filtered user profiles 
using a gazetteer that includes:
\begin{itemize}
    \item All Arab country names written in either Arabic, English, or French\footnote{French is widely used in the Maghreb region.} such as \<المغرب>
    (Almgrb -- Morocco), Morocco, and Maroc respectively.
    \item The names of major cities in these countries in both Arabic and English as specified in Wikipedia\footnote{\url{https://en.wikipedia.org/wiki/List_of_countries_by_largest_and_second_largest_cities}} such as \<القدس> 
    (Alqds -- Jerusalem) and \<وهران> 
    (whrAn -- Oran, Algeria).
    \item Arabic adjectives specifying all nationalities in both masculine and feminine forms with and without the definite article \<ال> (Al -- the) such as 
    \<عراقي> (ErAqy - Iraqi (m.)), \<عراقية> (ErAqyp - Iraqi (f.)), and 
    \<العراقي> (AlErAqy - the Iraqi (m.)).
\end{itemize}

\subsection{Arabic Variant Identification}
The second filter checks if the account mainly tweets in either dialectal Arabic or MSA.  Since Arabic users commonly switch between MSA and dialectal Arabic, and we were interested in strictly dialectal tweets, we sought to filter out MSA tweets. There are multiple ways to distinguish between dialectal and MSA text.  One such method involves using a list of strictly dialectal words ~\cite{Darwish:2014:verifiably}.  However, constructing such lists across multiple dialects can be challenging.  We opted to train a text classifier using a heuristically labeled tweet set.  Specifically, given 50 million tweets that we collected between March and September 2018, we assumed that tweets strictly containing the MSA relative pronouns \<الذي، الذى، التي، التى، الذين> 
("Al*y, Al*Y, Alty, AltY, Al*yn" - who/that in masculine, feminine and plural forms) were MSA, and those strictly containing the dialectal relative pronoun \<اللي، اللى>  
("Ally, AllY" -- who/that) were dialectal.  The major advantage of the dialectal relative pronoun \<اللي>  
is that it is present in most (if not all) Arabic dialects with the same meaning but not in MSA. Table \ref{tab:dialectAlly} shows some examples of such usage across different dialects.

\begin{table}[]
    \centering
    \begin{tabular}{l|r|l}
        Dialect & Example & Translation \\ \hline
        Egyptian & \<كنتي كويسه ايه اللي حصل> &  you were good, what happened\\
        Levantine & \<في ناس منيح اللي ما في متلن> & it's good there are no other people like them\\
        Gulf & \<ابي ابدل المكيف اللي في غرفتي> & I want to change the air conditioner in my room\\
        Maghrebi & \<اللي يحبها ما يديرلهاش المشاكل> 
        & he who loves her, should not cause her trouble 
        \\
    \end{tabular}
    \caption{Examples usages of dialectal relative pronoun across dialects.}
    \label{tab:dialectAlly}
\end{table}

In doing so, we extracted 3.09M MSA tweets and 3.17M dialectal tweets. For these tweets, we replaced user mentions with @USER, digits to NUM, emojis to EMOJI, URLs to URL, and the aforementioned relative pronouns with RELATIVE. In doing so, we eliminated Twitter-specific features, which are not linguistic in nature, and eliminated the effect of the relative pronouns we used to construct the dataset. 


We set aside 20k MSA and dialectal tweets for testing (10k for each).  We trained a fastText classifier~\cite{joulin2016bag}, which is a deep-learning-based classifier, using character n-grams ranging in length between 3 and 6 grams. We tested on the held-out test set, and the accuracy of distinguishing between MSA and dialectal Arabic was 98\%.  Using this classifier, we classified the tweets of the users.  We retained users, where at least 50\% of their tweets were dialectal.

\subsection{Appropriateness Identification}
The third filter removed users who were mostly tweeting vulgar, sexually explicit, or pornographic tweets.  To filter out these users, we used the obscene word list generated by Mubarak et al. \shortcite{mubarak2017abusive}, which contains 288 words and 127 hashtags.  We removed users if more than 50\% of their tweets contained vulgar words.

\subsection{Normalization}
Tweets often contain tokens that are specific to the Twitter platform such as hashtags and user mentions.  To improve generalization of trained models beyond tweets, 
we split hashtags into their semantic constituents~\cite{bansal2015towards,declerck2015processing}, and replaced user \@mentions and URLs with ``@USER'' and ``URL'' respectively. 

\subsection{Constructing the Dataset}
We sorted all the users by the number of their followers (in descending order) 
and retained the top 200 accounts from each country. 
After applying the three aforementioned filters, we ended up with 2,525 users from 18 countries (140 users per country on average), who authored 540k tweets (30k per country on average) with a total of 8.8M words.  
Table~\ref{tab:datasets_stats} provides per country breakdown of the dataset. 

\begin{table}[h]
    \centering
    \begin{tabular}{c|c|c|c|c|c|c|c|c|c}
Country & IQ & BH & KW & SA & AE & OM & QA & YE & SY \\ \hline
Users & 142 & 169 & 160 & 149 & 172 & 176 & 139 & 138 & 139 \\ \hline
Training Tweets (k) & 18.4 & 28.3 & 49.9 & 35.4 & 27.8 & 24.8 & 36.7 & 11.6 & 18.3 \\ \hline
Test tweets & 178 & 184 & 190 & 199 & 192 & 169 & 198 & 193 & 194 \\ \hline
\hline
Country & JO & PL & LB & EG & SD & LY & TN & DZ & MA \\ \hline
Users & 146 & 145 & 141 & 150 & 139 & 149 & 68 & 130 & 73 \\ \hline
Training Tweets (k) & 34.1 & 48.6 & 38.4 & 67.8 & 16.3 & 40.9 & 12.9 & 17.6 & 12.8 \\ \hline
Test tweets & 180 & 173 & 194 & 200 & 188 & 169 & 154 & 170 & 178 
    \end{tabular}
\caption{The number of users and tweets per country in our tweet corpus. 
}
\label{tab:datasets_stats}
\end{table}


\subsection{Data Validation}
To assess the quality of our new data set, we resorted to manual assessment, where we manually labeled a random sample of 200 tweets from the tweets of each country. Though some expressions may be unique to a dialect of a particular country (e.g. \<إزيك> ($<$zyk -- how are you (Egyptian))), other expressions may be used in dialects in different countries (e.g. \<لا باس> (lA bAs -- no problem or good (Algerian (DZ), Moroccan (MA), and Tunisian (TN)))).  Thus, the instruction we gave to the annotators was: ``Is this tweet consistent with the dialect spoken in your country?'' The labeling of the tweets from each country was done by native speakers from that country.  The average accuracy across countries was 91.5\%. 
When available for some countries, we asked a second annotator to also label the tweets.  For these countries, namely Egypt, Algeria, Saudi Arabia, and Syria, the average inter-annotator agreement was 87\%.  Incidentally, these four countries cover the major dialect groups.  

\begin{figure}[h]
\begin{center}
\includegraphics[scale=0.65]{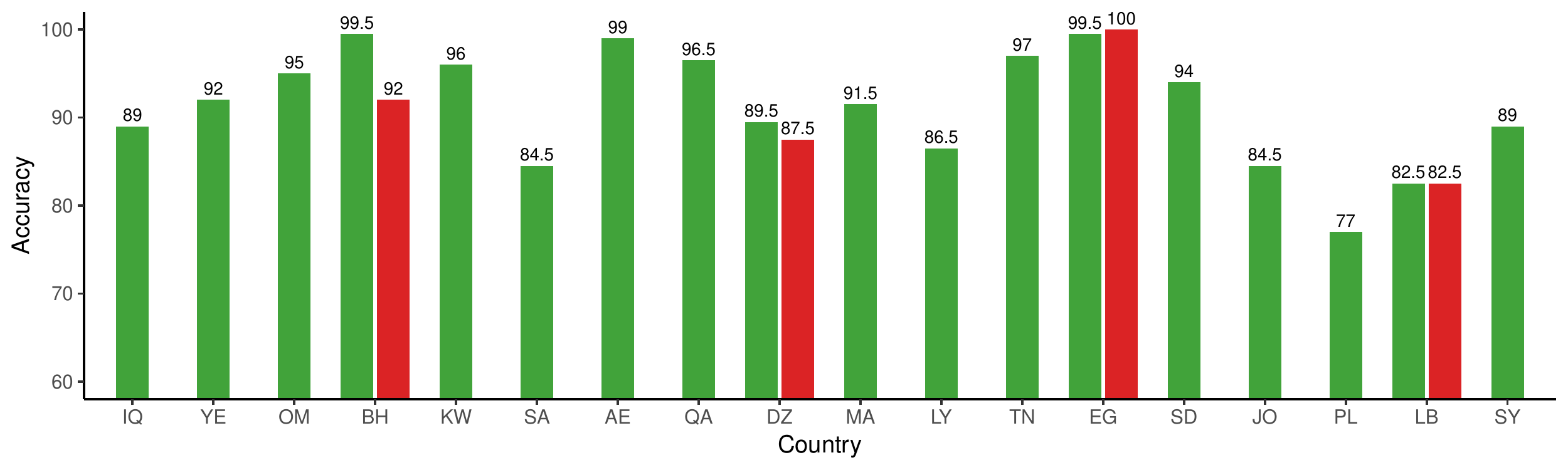} 
\caption{Annotation accuracy per country. Seconds annotators are colored in ``Red''. 
}
\label{fig.AnnoAccuracy} 
\end{center}
\end{figure}

Figure \ref{fig.AnnoAccuracy} shows the accuracy per country for all annotators. 
Of the 200 tweets per country, those that were judged as correctly labeled were removed from the dataset, and we used them as a test set. In all, we had 3,303 test tweets (with 183 tweets on average for each of the 18 countries).  Table~\ref{tab:datasets_stats} lists the number of test tweets per country.  
We are releasing the test set as a benchmark for dialect identification~\footnote{The test set can be downloaded from: \url{http://alt.qcri.org/resources/qadi/}}.

The manually rejected tweets that the annotators classified as not from their dialects were mostly cases where the users interacted with or responded to users from different countries. In such cases, users tend to code-switch or adopt the other users' dialect. For example, a user identified as Tunisian tweeted \<انا عموما بحب الظلمة اوي> (Ana EmwmA bHb AlZlmp Awy -- I generally like darkness a lot). The annotator correctly tagged this as not Tunisian (TN), as it is clearly Egyptian (EG). In this example, the Tunisian user was conversing with a person from Egypt or the Levant. In another example, the tweet \<هسه جاي تقول أحبك ؟>  (hsh jAy tqwl $>$Hbk -- just now you come to say I love you), the annotator labeled the tweet as not Yemeni (YE), mostly because of the typically Iraqi word ``\<هسه>'' (hsh -- just now). In this case, we found that the tweet was quoting a popular Iraqi song. 



\section{Corpus Statistics and Analysis}

Upon constructing the dataset, we attempted to explore its characteristics.  First, we extracted features that are distinctive for each dialect. To do so, we computed the so-called valence score for each word in each dialect \cite{conover2011political}. The score helps determine the distinctiveness of a given word in a specific dialect in reference to other dialects. Given $N(t,D_i)$, which is the frequency of the term $t$ in Dialect $D_i$, valence is computed as follows:
\begin{equation}
    V(t)_i = 2 \frac{
        \frac{N(t, D_i)}{N(D_i)}}
        { \sum_{n}\frac{N(t, D_n)}{N(D_n)}
    } - 1
\end{equation}

\begin{figure}[!h]
\begin{center}
\includegraphics[scale=0.55]{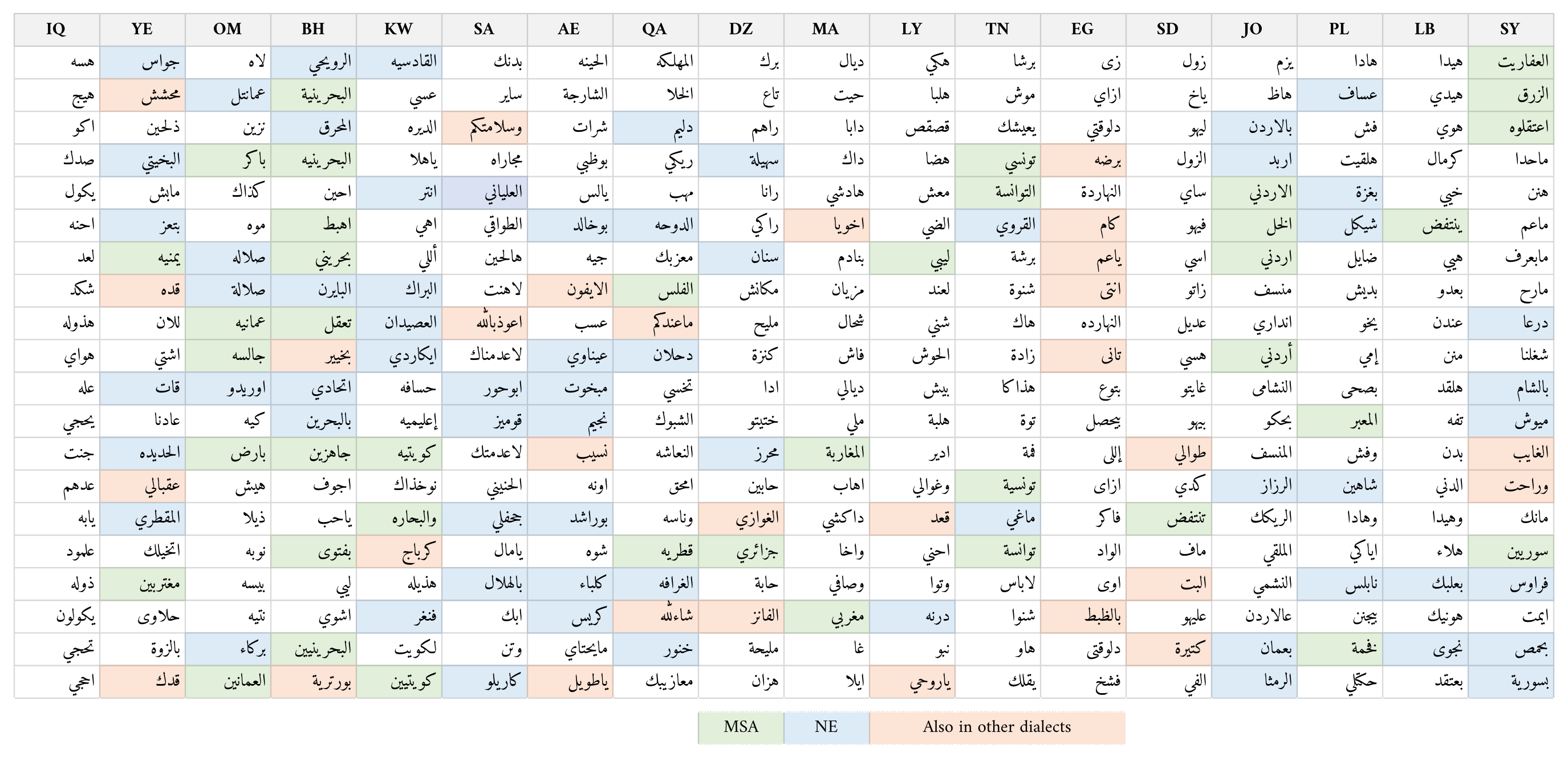} 
\caption{Highest valence words for each country.}
\label{fig.TopWords}
\end{center}
\end{figure}

Where $N(D_i)$ is the total number of occurrences of all words in the dialect $D_i$.  Figure~\ref{fig.TopWords} lists the words with highest valence scores per country.  Though the majority of the top words were in fact dialectal words that were distinctive for each country (typically function words), there were three other prominent categories of words that were not.  The first was names of locations inside these countries, which implies that geographic locations in a user's Twitter timeline can be a strong features in identifying the country of the user.  The second had words that appear in multiple dialects, which is expected given the overlap between dialects from different countries.  The third category included MSA words.  Though we intentionally excluded all tweets that were identified as MSA, the appearance of such words was expected given the large overlap between MSA and dialects and the frequent context switching between MSA and dialects in user tweets.


\begin{figure}[h]
\begin{center}
\includegraphics[scale=0.48]{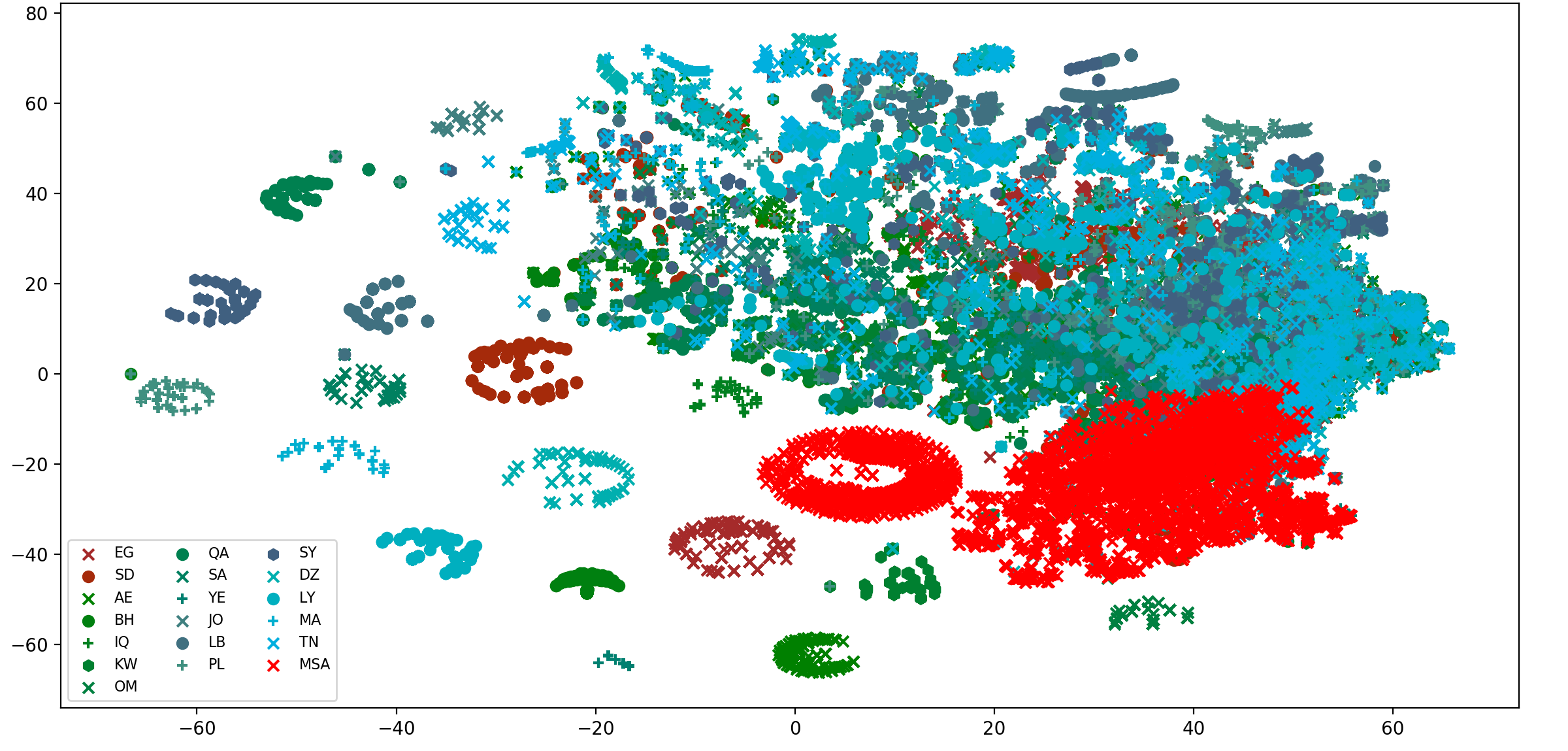} 
\caption{Arabic Dialects t-SNE projection using top 10k words with highest valence scores.}
\label{fig.1TSNEProject}
\end{center}
\end{figure}

Next, we computed the similarity between dialects to ascertain if similarities are consistent with reports in prior literature. To do so, we used two methods to visualize the similarity between different country-level dialects. In the first method, we constructed a list of the top 10k words with the highest valence scores across all dialects.  
The resulting list can be viewed as a vector of 19 valence values for each word corresponding to the valence of 18 different country-level dialects in addition to the MSA. For MSA data, we used the 3.09M MSA tweets that we used earlier to train the MSA/dialect classifier.  Then given the word vectors, we projected the words onto a two dimensional space using t-distributed Stochastic Neighbor Embedding (t-SNE) \cite{maaten2008visualizing}. Figure ~\ref{fig.1TSNEProject} shows the projection of the 10k words with the highest valence score. The projection illustrates that beyond the commonality and the overlap between the various dialects and MSA, we can still see islands that are unique to each dialect.  \newline

\begin{figure}[h]
\begin{center}
\includegraphics[scale=0.50]{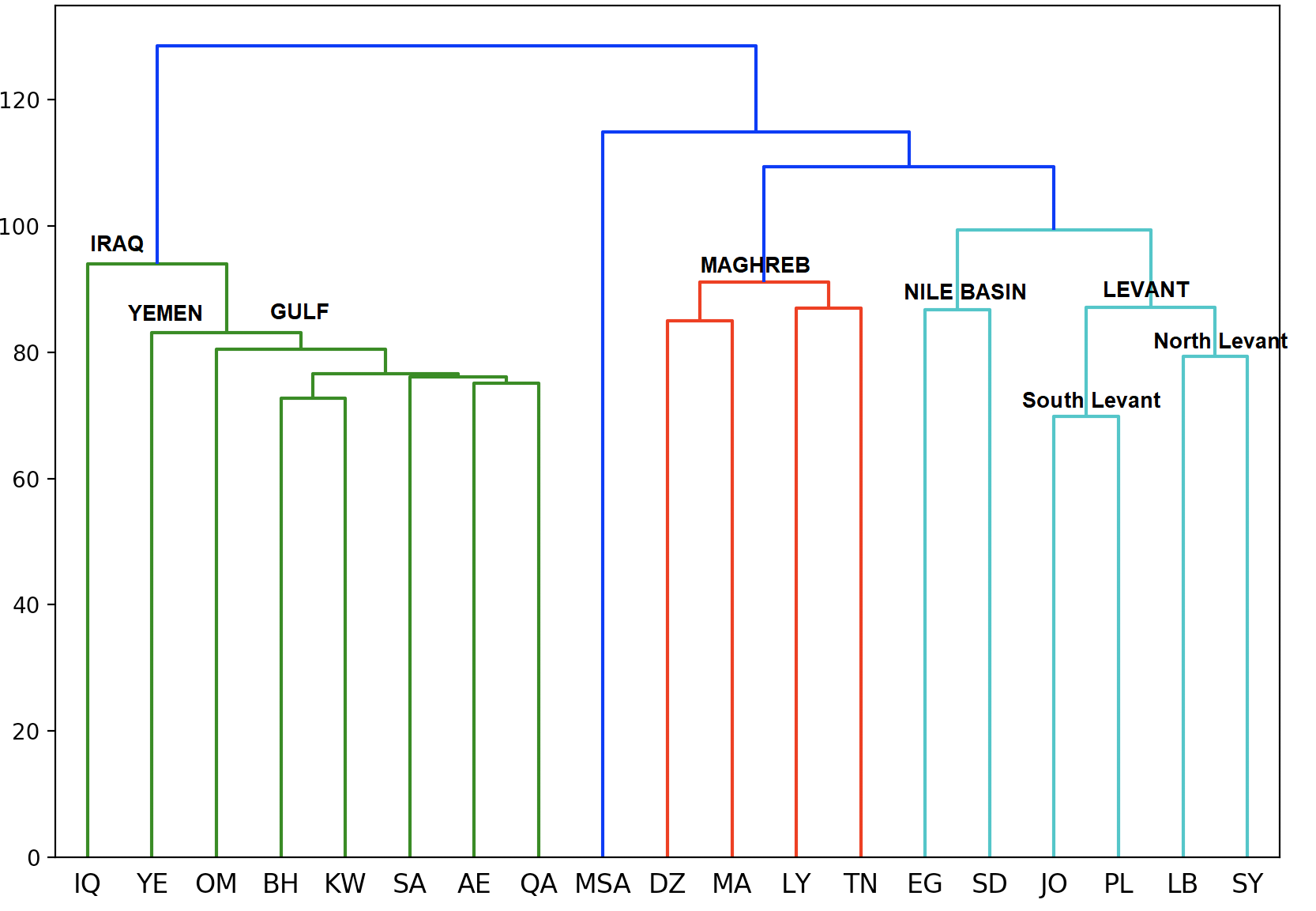} 
\caption{Clustering of Arabic Dialects using valence scores on top 10k words.}
\label{fig.HAC} 
\end{center}
\end{figure}

For the second visualization method, we used the previously computed valence scores for each of the top 10k words, and applied the SHC bottom-up hierarchical agglomerative clustering \cite{Li2009SHC}. The algorithm treats each dialect as a singleton cluster at the outset and then successively merges (or agglomerates) clusters until all clusters have been merged into a single cluster that contains all dialects. Figure~\ref{fig.HAC} shows the results of hierarchical clustering.  The figure reflects the similarity and the geographical proximity of various dialects. At higher levels, dialects are grouped per region, where we can identify the major dialectal groups, namely Gulf, Maghrebi, Egyptian, and Levantine. This is aligned with geographical distribution of the dialects as well as findings of other researchers~\cite{salameh-etal-2018-fine}.

\section{Experimental Setup}
Given our new dataset, we conducted a battery of experiments on the dataset to build effective country-level Arabic dialect identification.  We experimented with several tweet representation and classification models.  For tweet representations, we used: surface features, namely words and character n-grams, static embeddings, and deep contextual embeddings, namely AraBERT and BERT\textsubscript{base-multilingual}. For classification, we used two different classifiers, namely an SVM classifier and a fine-tuned BERT model.  For comparison, we conducted the same experiments on \MADAR dataset. 
In the following subsections, we present tweet representations and classification models.

\subsection{Representations}
\paragraph{Surface Features}  
We used two different surface-level features, namely word and character n-grams. Specifically, we represented tweets using: i) character n-grams, where we used 3 to 7-grams; ii) word n-grams, where we used 2 to 6-grams; 
and iii) a combination of word and character n-grams. For our dataset and \MADAR dataset, we normalized each instance of URL, number, and user mention to URL, NUM, and MENTION respectively.  

\paragraph{Static Embeddings}
We experimented with the \textbf{Mazajak} word-level skip-gram embeddings \cite{abu-farha-magdy-2019-mazajak}, which were trained on 250M Arabic tweets with 300-dimensional vectors.

\paragraph{Deep Contextualized Embeddings}
We also experimented with two pre-trained contextualized embeddings with fine-tuning for down-stream tasks, namely BERT\textsubscript{base-multilingual} (hereafter as simply mBERT) and AraBERT~\cite{Antoun2020AraBERT}. Recently, deep contextualized language models such as BERT (Bidirectional Encoder Representations from Transformers)~\cite{devlin-2019-bert}, UMLFIT~\cite{howard-ruder-2018}, and OpenAI GPT ~\cite{radford2018}, to name but a few, have achieved ground-breaking results in many NLP classification and language understanding tasks. 

Both mBERT and AraBERT are pre-trained on identical architectures, namely an encoder with $12$ Transformer blocks, hidden size of $768$, and $12$ self-attention heads.  However, they differ in one major way.  While mBERT is pre-trained on Wikipedia text for 104 languages, of which Arabic Wikipedia is a small portion, AraBERT is trained on a large Arabic news corpus containing 8.5M articles composed of roughly 2.5B tokens. AraBERT has two versions: one of which uses Arabic words that were segmented using Farasa~\cite{darwish2016farasa} and another one that uses SentencePiece (BP).  For consistency with mBERT, we used AraBERT with BP. Following \newcite{devlin-2019-bert}, the classification consists of introducing a dense layer over the final hidden state $h$ corresponding to first token of the sequence, [CLS], adding a softmax activation on the top of BERT to predict the probability  of the $l$ label:  $$p(l|h) = softmax(Wh),$$   where $W$ is the task-specific weight matrix. During fine-tuning, all mBERT or AraBERT parameters together with $W$ are optimized end-to-end to maximize the log-probability of the correct labels.

\subsection{Classification Models}
For classification, we used an SVM classifier and fine-tuned BERT and AraBERT.  We utilized the SVM classifier when using surface features and static pre-trained Mazajak embeddings.  We used the Scikit Learn libsvm implementations of the SVM classifier with a linear kernel.  When using contextualized embeddings, we fine-tuned BERT or AraBERT by adding a fully-connected dense layer followed by a softmax classifier, minimizing the binary cross-entropy loss function for the training data. For all experiments, we used the PyTorch\footnote{\url{https://pytorch.org/}} implementation by HuggingFace\footnote{\url{https://github.com/huggingface/transformers}} as it provides pre-trained weights and vocabulary.

\subsection{Experiments and Results}
As stated earlier, we ran a number of country-level dialect ID experiments on our new dataset and on \MADAR dataset for comparison. 
The details of the training and test splits for the dataset as as follows:

\paragraph{QADI Dataset}  Table~\ref{tab:datasets_stats} provides the statistics of the training and test parts of QADI dataset.  Given that manual verification was done at tweet-level, all the experiments on QADI dataset were done at tweet level.  In all, the dataset contains 540k training tweets and 3,303 test tweets.

\paragraph{\MADAR Dataset} 
\MADAR task 2 dataset was designed for user-level classification, where each user is assigned a country label.  The dataset is split into training, development, and test splits that contain 2,180, 
300, and 500 users respectively.  The dataset contains approximately 100 sample tweets for every users. For our experiments, we merged the training and development splits.  Since we were performing tweet-level classification, we assigned the user label to all the tweets of the user, and proceeded to perform tweet-level training and testing. We normalized the tweets in the same manner applied on the ~\QADI dataset, where we segmented hashtags and normalized user mentions and URLs to \@USER and URL respectively. 


\subsubsection{Results}  
\begin{figure*}[h]
\begin{center}
\includegraphics[scale=0.45]{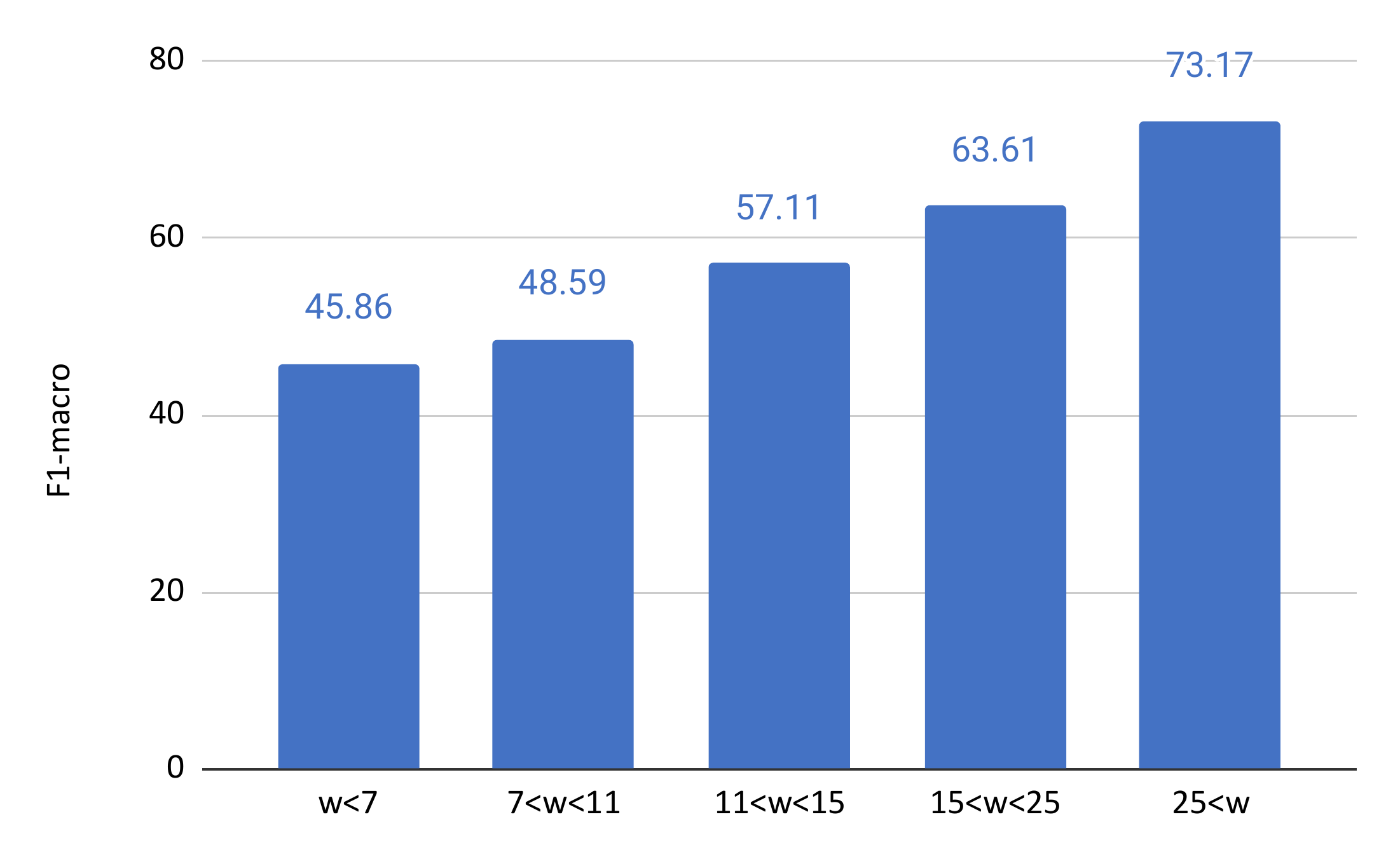} 
\caption{Macro-averaged F1-score given tweet length using AraBERT.}
\label{f1_accuracy_length}
\end{center}
\end{figure*}

Table~\ref{tab:QADI-MADAR-eval-results} reports on the results of training and testing using \QADI and \MADAR datasets.  As \QADI results show, using contextual embeddings yielded the best results with AraBERT results edging mBERT results.  Using an SVM classifier that is trained using a combination of character and word n-grams (CW26) was slightly lower than using contextual embeddings.  However, this setup is computationally more efficient than using contextual embeddings. Using Mazajak embeddings led to significantly lower results.  Further, when inspecting the best classification results (AraBERT), we noted that the length of the tweets impacted the classification results. The longer a tweet, the more accurate the prediction was. Figure~\ref{f1_accuracy_length} shows the accuracy of the classifier for various tweet lengths. This is expected given that longer tweets potentially contain more clues for the classifier. 

Training using \MADAR led to significantly lower results compared to training using \QADI.  This likely stems from a mismatch between the problem at hand (tweet-level dialect ID) and the purpose for which \MADAR was constructed (user-level dialect/country ID). Further, belonging to a country does not guarantee that a user will always tweet in the dialect of that country. Often users from different countries use MSA (or even other languages). We speculated that many of the tweets in the MADAR data are actually MSA, because the tweets were collected without taking into account whether they were actually dialectal or not. To test this hypothesis, we used our aforementioned MSA/dialectal classifier. When we classified the MADAR tweets, the classifier tagged 29\% of the tweets as dialectal and the rest (71\%) as MSA -- confirming our hypothesis. Since the vast majority of the tweets were MSA, training on the MADAR dataset led to significantly lower tweet-level dialect classification results. Since QADI filters out MSA tweets, it doesn't have the same issue.

\begin{table}[h]
\centering
\begin{tabular}{lcc}
            & \multicolumn{2}{c}{Training Set} \\
Classifier & QADI & MADAR \\ \hline
MultiLangBERT    & 58.9 & 25.3  \\
AraBERT     & \textbf{60.6} & 29.0 \\
Mazajak     & 39.8 & 24.6 \\
SVM(CW26)   & 57.2 & 24.0 \\
SVM(C37)    & 48.1 & 21.2 \\ \hline
\end{tabular}
\caption{Classification results for \MADAR and \QADI test sets using the various models}
\label{tab:QADI-MADAR-eval-results}
\end{table}

\begin{figure*}[h!]
\begin{center}
\includegraphics[scale=0.75]{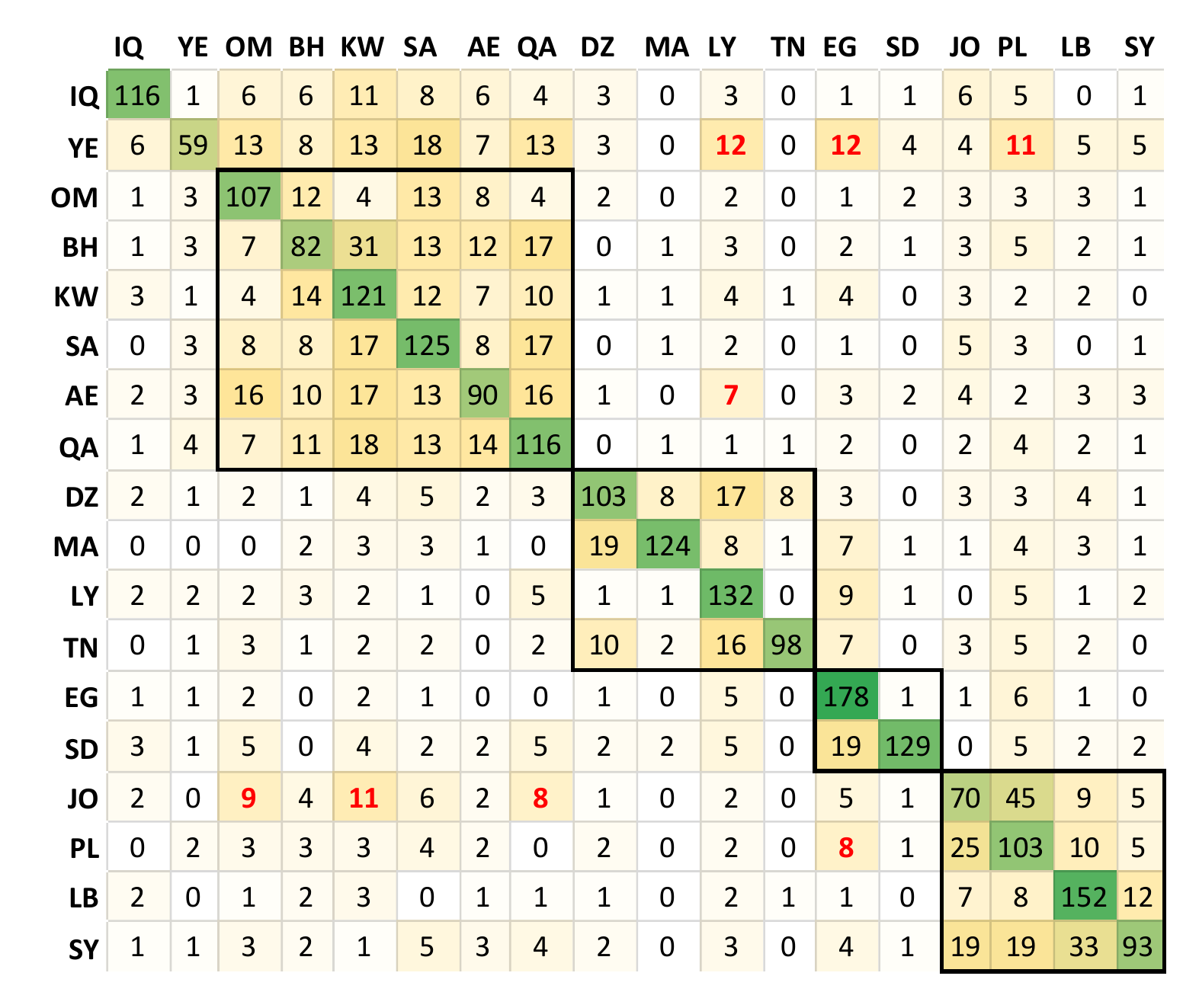} 
\caption{Confusion matrix for the test dataset. The bulk of the mis-classification happens within the regions (marked withs thick border). Outliers are marked in \textcolor{red}{red} where the classification is beyond the region. }
\label{CM}
\end{center}
\end{figure*}

\subsubsection{Error Analysis}
We inspected tweets from the QADI test set that were misclassified by AraBERT, as this setup yielded the best results. Generally, the most prominent reason for incorrect classification could be attributed to the fluidity of geolinguistic distinctions between Arabic dialects. To some degree, geographical proximity is contingent with dialectal closeness, making it difficult for the system to distinguish between the dialects at hand. Note that these dialects share a plethora of linguistic features to warrant their subsumability under the same dialect. As shown in Figure~\ref{CM}, the dialects from the Gulf region (OM, BH, KW, SA, AE, and QA) depict the largest confusion as they get mixed up among themselves.  For example, the tweet, \< ...دخلت البرنامج سويت حساب والرقم ما يدخل وكلمتكم كم مره   
شوفو لي حل> 
(I entered the program, created an account, and the number is not accepted ... I called you many times. Find me a solution), could be plausibly attributed to any of the Gulf dialects. 
Similarly, the second largest confusion is among dialects from the Levant region (JO, PL, LB, and SY), where we found a considerable amount of the mix-up between LB and SY. The tweet, \<لما الانسان بفكر حالو الوحيد يلي بيفهم . . تأكد ان حمار> 
(When the human thinks that he is the only one who understands ... be sure that he is a donkey), can be equally valid for both dialects. Similar to the results observed for both the Gulf and Levant regions, the Maghrebi dialects (MA, DZ, LY, TN) exhibit a similar pattern. MA and DZ account for considerable confusion. For instance, the tweet \<الله يبارك فيك خويااا> 
(God bless you, brother!!), could be used in both dialects. As for the Nile Basin dialects, Egyptian (EG) and Sudanese (SD) could also be confused with one another. The tweet, \<التويته دي معدلة فوتوشوب>  
(This tweet
is modified in Photoshop), is equally valid in both dialects. This is normal since SD is particularly similar to central and southern Sa'idi Egyptian Arabic.\footnote{\url{https://en.wikipedia.org/wiki/Egyptian_Arabic}} Interestingly, we found that about  $2\%$ of the misclassified tweets were outliers-classified outside of their region, highlighted in red in Figure~\ref{CM}. The main reasons for incorrect classification, beyond the region, is due to the fact that many of them contain quotes from popular songs and poems or in few cases they have MSA words. As this YE tweet, \< وإنت عارف مكانتك ومتأكد منها من غير مايعيشك شعور مرات إنت العمر ومرات ما أعرفك ...>
(And you know your status, and you are certain about it without making you feel...), misclassified as LY. This tweet despite being manually labelled as YE, it could fit in other countries.
\section{Conclusion}
In this paper, we presented a method for building country-level dialectal tweet corpus.  
The construction of the corpus relied on a cascade of filters, where user accounts were filtered on keywords that were indicative of country, and tweets were filtered to remove users who predominantly tweet in MSA or vulgar language.  We used our method to build a large corpus containing 540k tweets from 2,525 Twitter accounts that cover 18 Arab countries. Based on a manual inspection of a random sample of tweets from the corpus, the estimated accuracy of country-level dialectal tags was 91.5\%.  We also showed that the resultant corpus can be effective in training a country-level dialect classifier for tweets that achieves a macro-averaged F1-score of 60.6\% across 18 different classes.  We compared to training on a publicly available dataset, namely \MADAR dataset, and \MADAR results were significantly lower.  

Based on our error analysis, we discovered that a large source of errors was due to the naturally occurring overlap between dialects from neighboring countries and to code switching between different dialects. Code switching happens when users adopt another dialect when they communicate with users from other countries, or when they use more than one dialect in the same tweet.  While overlap between dialects is potentially an intractable problem, detecting code switching between dialects is a future direction that can further help filter training data and identify tweets that may include multiple dialects simultaneously. For future work, we plan to investigate code switching. We also plan to examine the efficacy of extending our dialectal dataset to performing user-level geotagging classification. Though identifying a user's country may depend on multiple signals, accurate dialect identification is likely a strong signal that can aid classification.  

\bibliographystyle{coling}
\bibliography{coling2020}

\end{document}